\documentclass{article}

\PassOptionsToPackage{numbers, compress}{natbib}
\usepackage[final]{neurips_2021}

\usepackage[utf8]{inputenc} % allow utf-8 input
\usepackage[T1]{fontenc}    % use 8-bit T1 fonts
\usepackage{hyperref}       % hyperlinks
\usepackage{url}            % simple URL typesetting
\usepackage{booktabs}       % professional-quality tables
\usepackage{amsfonts}  
\usepackage{amsmath}
\usepackage{amsthm}
\usepackage{bm}

\usepackage[dvipsnames]{xcolor}

\usepackage{wrapfig}
\usepackage{nicefrac}       % compact symbols for 1/2, etc.
\usepackage{microtype}      % microtypography
\usepackage{xcolor}         % colors
\usepackage{graphicx}       % figures
\usepackage{subcaption}     % for subfigures
\usepackage{paralist}
\usepackage{microtype}
\usepackage{amssymb}
\usepackage{pdfpages}
\usepackage{fancyhdr}
\usepackage{graphicx}
\usepackage{caption}
\usepackage{subcaption}
\usepackage{multirow}
\usepackage{nicefrac}       % compact symbols for 1/2, etc.
\usepackage{stmaryrd}
\usepackage[parfill]{parskip}
\usepackage{mathtools}
\usepackage{listings}
\usepackage{cases}
\usepackage{tabularx}
\usepackage{comment}
\usepackage{paralist}
\usepackage{breqn}
\usepackage{color}
\usepackage{algorithm, algorithmic}
\usepackage{appendix}
\usepackage{xspace}
\usepackage{enumitem}
\usepackage{gensymb}
\usepackage[english]{babel}
\usepackage{xcolor}
\usepackage{todonotes}
\usepackage{wrapfig}
\usepackage{letltxmacro}
\usepackage{breqn}
\usetikzlibrary{shapes.geometric,arrows,arrows.meta, fit}
\mathchardef\mhyphen="2D

\newcommand{\vertiii}[1]{{\left\vert\kern-0.25ex\left\vert\kern-0.25ex\left\vert #1
    \right\vert\kern-0.25ex\right\vert\kern-0.25ex\right\vert}}

\definecolor{codegreen}{rgb}{0,0.6,0}
\definecolor{codegray}{rgb}{0.5,0.5,0.5}
\definecolor{codepurple}{rgb}{0.58,0,0.82}
\definecolor{backcolour}{rgb}{0.95,0.95,0.92}

\lstdefinestyle{mystyle}{
  backgroundcolor=\color{backcolour},   commentstyle=\color{codegreen},
  keywordstyle=\color{magenta},
  numberstyle=\tiny\color{codegray},
  stringstyle=\color{codepurple},
  basicstyle=\ttfamily\footnotesize,
  breakatwhitespace=false,         
  breaklines=true,                 
  captionpos=b,                    
  keepspaces=true,                 
  numbers=left,                    
  numbersep=5pt,                  
  showspaces=false,                
  showstringspaces=false,
  showtabs=false,                  
  tabsize=2
}

\hypersetup{
colorlinks = true,
linkcolor = ForestGreen,
anchorcolor = blue,
citecolor = ForestGreen,
filecolor = cyan,
menucolor = ForestGreen,
runcolor = cyan,
urlcolor = ForestGreen}

\usepackage{todonotes}
\presetkeys{todonotes}{inline}{}
\usepackage[normalem]{ulem}

\newcommand{\CF}[0]{\check{x}}

\newcommand{\F}[0]{x}

\newcommand*\samethanks[1][\value{footnote}]{\footnotemark[#1]}

\DeclareMathOperator*{\argmin}{arg\,min}

\title{CARLA: A Python Library to Benchmark Algorithmic Recourse and Counterfactual Explanation Algorithms}

\author{%
  Martin Pawelczyk\thanks{Corresponding author} \\
  University of Tübingen \\
  \texttt{martin.pawelczyk@uni-tuebingen.de} \\
  \And
    Sascha Bielawski \\
  University of Tübingen \\
  \texttt{sascha.bielawski@uni-tuebingen.de} \\
  \And
    Johannes van den Heuvel \\
  University of Tübingen \\
  \texttt{johannes.van-den-heuvel@uni-tuebingen.de} \\
  \And
    Tobias Richter \thanks{Equal senior author contribution} \\
  CarePay International \\
  \texttt{t.richter@carepay.com} \\
  \And
    Gjergji Kasneci \samethanks[2]{} \\
  University of Tübingen \\
  \texttt{gjergji.kasneci@uni-tuebingen.de} \\
}

\begin{document}

\maketitle

\begin{abstract}
Counterfactual explanations provide means for prescriptive model explanations by suggesting actionable feature changes (e.g., increase income) that allow individuals to achieve favourable outcomes in the future (e.g., insurance approval).
Choosing an appropriate method is a crucial aspect for meaningful counterfactual explanations. As documented in recent reviews, %e.g.~\cite{karimi2020survey,verma2020counterfactual}, 
there exists a quickly growing literature with available methods. Yet, in the absence of widely available open--source implementations, the decision in favour of certain models is primarily based on what is readily available. Going forward -- to guarantee meaningful comparisons across explanation methods -- we present \texttt{CARLA} (\textbf{C}ounterfactual \textbf{A}nd \textbf{R}ecourse \textbf{L}ibr\textbf{A}ry), a python library for benchmarking counterfactual explanation methods across both different data sets and different machine learning models. In summary, our work provides the following contributions: (i) an extensive benchmark of 11 popular counterfactual explanation methods, (ii) a benchmarking framework for research on future counterfactual explanation methods, and (iii) a standardized set of integrated evaluation measures and data sets for transparent and extensive comparisons of these methods.
We have open sourced \texttt{CARLA} and our experimental results on \href{https://github.com/indyfree/CARLA}{Github}, making them available as competitive baselines. We welcome contributions from other research groups and practitioners.
\end{abstract}

\section{Introduction} \label{section:intro}
Machine learning (ML) methods have found their way into numerous everyday applications and have become an indispensable asset in various sensitive domains, like disease diagnostics \cite{fatima2017survey}, criminal justice \cite{berk2012criminal}, or credit risk scoring \cite{khandani20102767}. While ML models bear the great potential to provide effective support in human decision making processes, their predictions may have considerable impact on personal lives, where the final decisions might be disadvantageous for an end user. For example, the rejection of a loan or the denial of parole might have negative effects on the future development of the corresponding person's life. 

When ML systems involve humans in the loop, it is crucial to build a strong foundation for long-term acceptance of these methods. To this end, it is critical (1) to \emph{explain} the predictions of a model and (2) to \emph{offer constructive means for the improvement} of those predictions to the advantage of the end--user. Counterfactual explanations
%\footnote{The terms counterfactual explanations~\citep{wachter2017counterfactual}, contrastive explanations~\citep{karimi2020survey}, and algorithmic recourse~\citep{ustun2019actionable} have been used interchangeably in prior work. We use these terms interchangeably to refer to the notion introduced by \citet{wachter2017counterfactual}.} 
-- popularized by the seminal work of \cite{wachter2017counterfactual} -- provide means for prescriptive model explanations by suggesting actionable feature changes (e.g., increase income) that allow individuals to achieve favourable outcomes in the future (e.g., insurance approval).

\begin{wrapfigure}[21]{r}{0.42\textwidth}
    \centering
    \includegraphics[scale=0.5]{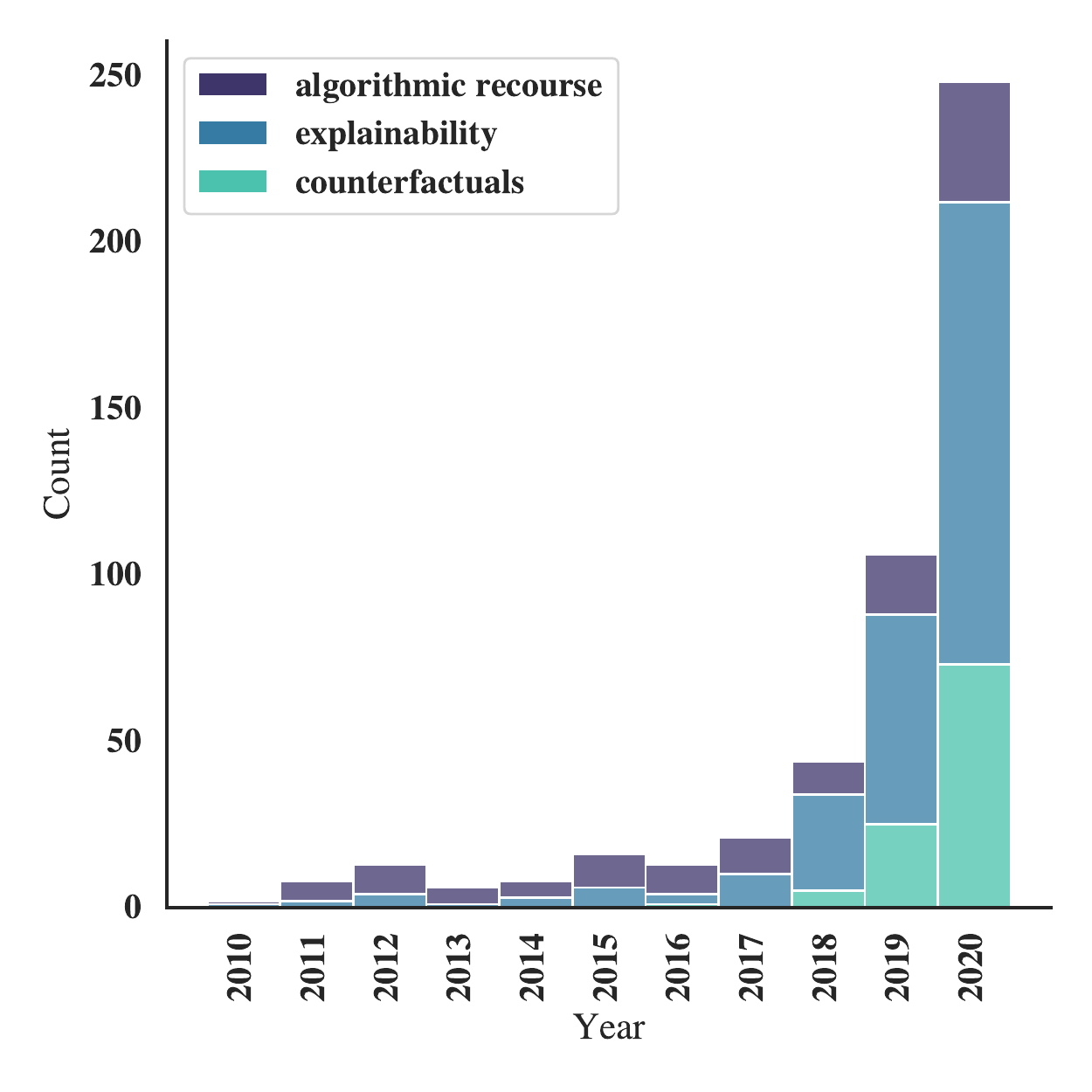}
    \caption{ArXiv submissions over time on explainability, counterfactual explanations and algorithmic recourse.}
    \label{fig:growth_explainability_papers}
\end{wrapfigure}

When counterfactual explainability is employed in systems that involve humans in the loop,  the community refers to it as \emph{recourse}. Algorithmic recourse subsumes precise recipes on how to obtain desirable outcomes after being subjected to an automated decision, emphasizing feasibility constraints that have to be taken into account. Those explanations are found by making the smallest possible change to an input vector to influence the prediction of a pretrained classifier in a positive way; for example, from `loan denial' to `loan approval’, subject to the constraint that an individual's \texttt{sex} may not change. As documented in recent reviews, 
there exists a quickly growing literature with available methods (see Figure \ref{fig:growth_explainability_papers} and \citep{stepin2021survey,karimi2020survey,verma2020counterfactual}), reflecting the insight that the understanding of complex machine learning models is an elementary ingredient for a wide and safe technology adoption.

In practice, the counterfactual explanation (CE) that an individual receives crucially depends on the method that computes the recourse suggestions. Hence, there is a substantial need for a standardized benchmarking platform, which ensures that methods can be compared in a transparent and meaningful way. Researchers need to be able to easily evaluate their proposed methods against the overwhelming diversity of already available methods and practitioners need to make sure that they are using the right recourse mechanism for the problem at hand. Therefore, a standardized framework for comparison and quality assurance is an essential and indispensable prerequisite.

In this work, we present \texttt{CARLA} (\textbf{C}ounterfactual \textbf{A}nd \textbf{R}ecourse \textbf{L}ibr\textbf{A}ry), a python library with the following merits:
First, \texttt{CARLA} provides \textbf{competitive baselines} for researchers to benchmark \textit{new} counterfactual explanation and recourse methods for the standardized and transparent comparison of CE methods on different integrated data sets.
Second, \texttt{CARLA} is a \textbf{common framework} with more than 10 counterfactual explanation methods in combination with the possibility to easily integrate new methods into a commonly accessible and easily distributable Python library. Moreover, the built-in integrated evaluation measures allow users to plug-in their custom black-box predictive models into the available counterfactual explanation methods and conduct extensive evaluations in comparison with other recourse mechanisms across different data sets. The same is true for researchers, who can use \texttt{CARLA} to extensively benchmark available counterfactual methods on popular data sets across various ML models. 
Third, \texttt{CARLA} \textbf{supports popular optimization frameworks} such as Tensorflow \cite{abadi2016tensorflow} and PyTorch \cite{paszke2019pytorch}, and provides a generic abstraction layer to support custom implementations. Users can can define problem--specific data set characteristics like immutable features and explicitly specify hyperparameters for the chosen counterfactual explanation method.

The remainder of this work is structured as follows: Section \ref{section:related_work} presents related work, Section \ref{section:background} formally introduces the recourse problem, Section \ref{section:benchmarking_process} presents the benchmarking process. In Section~\ref{section:evaluation}, we describe our main findings, before concluding in Section~\ref{section:conclusion}. Appendices \ref{appendix:carla_software} - \ref{appendix:hyperparameters} describe \texttt{CARLA}'s software architecture and usage instructions, as well as additional experimental results, used ML classifiers, data sets and hyperparameters settings.
%\todo[author=Tobi]{@Martin: Check if introducing the appendix works like I did above}

\section{Related Work}\label{section:related_work}

Explainable machine learning is concerned with the problem of providing explanations for complex ML models. Towards this goal, various streams of research follow different explainability paradigms which can be categorized into the following groups \cite{guidotti2018survey,gade2019explainable}.

\subsection{Feature Highlighting Explanations}
 \textbf{Local input attribution techniques} seek to explain the behaviour of ML models instance by instance. Those methods aim to understand how all inputs available to the model are being used to arrive at a certain prediction. Some popular approaches for model explanations aim at explainability by design \cite{lou2012intelligible,alvarez2018towards,broelemann2018gradient,wang2019designing}. For white-box models -- the internal model parameters are known -- gradient-based approaches, e.g.~\cite{kasneci2016licon,chattopadhay2018grad} (for deep neural networks), and rule-based or probabilistic approaches for tree ensembles, e.g.~\cite{hara2018making,deng2019interpreting} have been proposed. In cases where the parameters of the complex models cannot be accessed, model-agnostic approaches can prove useful. This group of approaches seeks to explain a model's behavior locally by applying surrogate models~\cite{ribeiro2016should, lundberg2017unified,ribeiro2018anchors,lundberg2020local}, which are interpretable by design and are used to explain
individual predictions of black-box ML models.
%In contrast to model explanations, such local explanations do not have to be globally valid.

% \paragraph{\textbf{Explanations by example}} Example-based explanation methods select particular instances from the dataset or exploit particular instances provided by human experts to explain the behavior of an ML model or to explain the underlying data distribution~\cite{mittelstadt2019explaining, gade2019explainable}.

%\paragraph{\textbf{Libraries}}

\subsection{Counterfactual Explanations}

The main purpose of counterfactual explanations is to suggest constructive interventions to the input of a complex model so that the output changes to the advantage of an end user. By emphasizing both the feature importance and the recommendation aspect, counterfactual explanation methods can be further divided into three different groups: independence-based, dependence-based, and causality-based approaches.

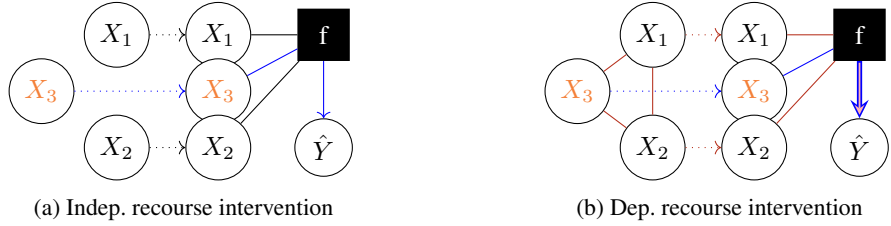
\begin{figure}[tb]
     \centering
     \begin{subfigure}[b]{0.49\columnwidth}
         \centering
         %\tikzstyle{line} = [draw,latex']

\begin{tikzpicture}[square/.style={regular polygon,regular polygon sides=4}]

\usetikzlibrary{fit}

\tikzstyle{circle1} = [circle, minimum size = 7mm, text centered, draw=black, fill=red!0]
\tikzstyle{circle2} = [circle, minimum size = 7mm, text centered, draw=black, fill=blue!20]

\node(a)[circle1] {$X_1$};
\node(b)[circle1, xshift=+0.0cm, yshift=-1.5cm] {$X_2$};
\node(c)[circle1, xshift=-1cm, yshift=-0.75cm] {$\textcolor{Orange}{X_3}$};
% \node(d)[circle1,  xshift=-1.0cm,  yshift=-2.25cm] {$\cdots$};
% \node(e)[circle1, yshift=-3cm] {$X_D$};

\node(a2)[circle1, xshift=+1.35cm] {$X_1$};
\node(c2)[circle1, xshift=+1.35cm, yshift=-0.75cm] {$\textcolor{Orange}{X_3}$};
\node(b2)[circle1, xshift=+1.35cm, yshift=-1.5cm] {$X_2$};
% \node(d2)[circle1, xshift=+2cm, yshift=-2.25cm] {$\cdots$};
% \node(e2)[circle1, xshift=+2cm, yshift=-3cm] {$X_D$};

%\node[draw, orange, fit=(a2) (b2) (c2) (d2) (e2)] {};

%\draw[-]  (a) -- (b);
%\draw[-]  (b) -- (c);
%\draw[-]  (a) -- (c);
%\draw[-]  (d) -- (e);

\draw[dotted,->]  (a) -- (a2);
\draw[dotted,->]  (b) -- (b2);
\draw[dotted, blue,->]  (c) -- (c2);
% \draw[dotted,->]  (d) -- (d2);
% \draw[dotted,->]  (e) -- (e2);

\node (f) [square, fill=black, draw, thick, xshift=+2.75cm, yshift=0cm] {\textcolor{white}{f}};
\node (y) [circle1, xshift=+2.75cm, yshift=-1.5cm] {$\hat{Y}$};

\draw[-]   (f) -- (a2);
\draw[-]   (f) -- (b2);
\draw[-, blue]   (f) -- (c2);
% \draw[-]   (f) -- (d2);
% \draw[-]   (f) -- (e2);

\draw[->,blue]   (f) -- (y);

\end{tikzpicture}
         \caption{Indep.\ recourse intervention}
         \label{fig:world_independent}
     \end{subfigure}
     \hfill
     \begin{subfigure}[b]{0.49\columnwidth}
         \centering
         %\tikzstyle{line} = [draw,latex']

\begin{tikzpicture}[square/.style={regular polygon,regular polygon sides=4}]

\usetikzlibrary{fit,arrows,decorations.markings}

\tikzstyle{circle1} = [circle, minimum size = 7mm, text centered, draw=black, fill=red!0]
\tikzstyle{circle2} = [circle, minimum size = 7mm, text centered, draw=black, fill=blue!20]

\tikzset{
  double arrow/.style args={#1 colored by #2 and #3}{
    -stealth,line width=#1,#2, % first arrow
    postaction={draw,-stealth,#3,line width=(#1)/3,
                shorten <=(#1)/3,shorten >=2*(#1)/3}, % second arrow
  }
}

%\tikzstyle{line} = [draw,latex']

\node(a)[circle1] {$X_1$};
\node(b)[circle1, xshift=+0.0cm, yshift=-1.5cm] {$X_2$};
\node(c)[circle1, xshift=-1.0cm, yshift=-0.75cm] {$\textcolor{Orange}{X_3}$};
% \node(d)[circle1,  xshift=-1.0cm,  yshift=-2.25cm] {$\cdots$};
% \node(e)[circle1, yshift=-3cm] {$X_D$};

\node(a2)[circle1, xshift=+1.35cm] {$X_1$};
\node(c2)[circle1, xshift=+1.35cm, yshift=-0.75cm] {$\textcolor{Orange}{X_3}$};
\node(b2)[circle1, xshift=+1.35cm, yshift=-1.5cm] {$X_2$};
% \node(d2)[circle1, xshift=+2cm, yshift=-2.25cm] {$\cdots$};
% \node(e2)[circle1, xshift=+2cm, yshift=-3cm] {$X_D$};

\draw[-, BrickRed]  (a) -- (b);
\draw[-, BrickRed]  (b) -- (c);
\draw[-, BrickRed]  (a) -- (c);
% \draw[-]  (d) -- (e);

\draw[dotted,BrickRed,->]  (a) -- (a2);
\draw[dotted,BrickRed,->]  (b) -- (b2);
\draw[dotted,blue,->]  (c) -- (c2);
% \draw[dotted,->]  (d) -- (d2);
% \draw[dotted,->]  (e) -- (e2);

\node (f) [square, fill=black, draw, thick, xshift=+2.75cm, yshift=0cm] {\textcolor{white}{f}};
\node (y) [circle1, xshift=+2.75cm, yshift=-1.5cm] {$\hat{Y}$};

\draw[-,BrickRed]   (f) -- (a2);
\draw[-,BrickRed]   (f) -- (b2);
\draw[-,blue]   (f) -- (c2);
% \draw[-]   (f) -- (d2);
% \draw[-]   (f) -- (e2);

\draw[double arrow=2pt colored by blue and red!30!white]   (f) -- (y);

\end{tikzpicture}
         \caption{Dep.\ recourse intervention }
         \label{fig:world_dependent}
     \end{subfigure}
     \caption{Different views on recourse generation. In (\subref{fig:world_independent}) a change to \textcolor{Orange}{$X_3$} only impacts $f$ through \textcolor{Orange}{$X_3$} , while in (\subref{fig:world_dependent}) the same change induces \textcolor{BrickRed}{indirect} effects on $f$, if \textcolor{Orange}{$X_3$} is correlated with other inputs.}
     \label{fig:schematic_comparison}
\end{figure}

In the class of \textbf{independence-based methods}, where the input features of the predictive model are assumed to be independent, some approaches use combinatorial solvers or evolutionary algorithms to generate recourse in the presence of feasibility constraints  \cite{ustun2019actionable,russell2019efficient,rawal2020individualized,karimi2019model,kenny2020generating,dandl2020multi}. Notable exceptions from this line of work are proposed by \cite{tolomei2017interpretable,laugel2017inverse,lash2017generalized,gupta2019equalizing,ghazimatin2020prince}, who use decision trees, random search, support vector machines (SVM) and information networks that are aligned with the recourse objective. Another line of research deploys gradient-based optimization to find low-cost counterfactual explanations in the presence of feasibility and diversity constraints \cite{Dhurandhar2018,mittelstadt2019explaining,mothilal2020fat,schut2021generating,van2019interpretable,pawelczyk2021connections}. The main problem with these approaches is that they abstract from input correlations. That implies that the intervention costs (i.e., the costs of changing the input to achieve the proposed counterfactual state) are too optimistically estimated. In other words, the estimated costs do not reflect the true costs that an individual would need to incur in practical scenarios, where feature dependencies are usually present: e.g., \emph{income} is dependent on \emph{tenure}, and if \emph{income} changes, \emph{tenure} also changes (see Figure \ref{fig:schematic_comparison} for a schematic comparison).

In the class of \textbf{causality-based} approaches, all methods make use of Pearl's causal modelling framework \citep{pearl2009causality}. As such, they usually require knowledge of the system of causal structural equations \citep{joshi2019towards,goyal2019explaining,karimi2020intervention,oshaughnessy2020generative} or the causal graph \citep{karimi2020probabilistic}. The authors of \cite{karimi2020intervention} show that these models can generate minimum-cost recourse, if the access to the true causal data generating process was available. However, in practical scenarios, the guarantee for such minimum-cost recommendations is vacuous, since, in complex settings, the causal model is likely to be miss-specified \citep{karimi2020probabilistic}. Since these methods usually require the \emph{true} causal graph -- which is the limiting factor in practice -- we have not considered them at this point, but we plan to do that in the future. 

\textbf{Dependence-based methods} bridge the gap between the strong independence assumption and the strong causal assumption. This class of models builds recourse suggestions on generative models  \cite{pawelczyk2019,downs2020interpretable,joshi2019towards,mahajan2019preserving,pmlr-v124-pawelczyk20a}. The main idea is to change the geometry of the intervention space to a lower dimensional latent space, which encodes different factors of variation while capturing input dependencies. To this end, these methods primarily use variational autoencoders (VAE) \citep{kingma2013auto,nazabal2018handling}. In particular, \citet{mahajan2019preserving} demonstrate how to encode various feasibility constraints into VAE-based models. Most recently, \cite{antoran2020getting} proposed \texttt{CLUE}, a generative recourse model that takes a classifier's uncertainty into account. Work that deviates from this line of research was done by \cite{Poyiadzi2020,Kanamori2020}. The authors of \cite{Poyiadzi2020} provide \texttt{FACE}, which uses a shortest path algorithm on graphs to find counterfactual explanations. In contrast, \citet{Kanamori2020} use integer programming techniques to account for input dependencies.

\section{Preliminaries} \label{section:background}
In this Section, we review the algorithmic recourse problem and draw a distinction between two observational (i.e., non--causal) methods.

\subsection{Counterfactual Explanations for Independent Inputs} \label{section:background_recourse_independent}
%\todo[author=Klaus,inline]{Maybe sharpen a bit, $y$ (discrete)  is not the output of $f$ (continuous).}
Let $\mathcal{D}$ be the data set consisting of $N$ input data points, $\mathcal{D} = \{(x_1, y_1), \dots, (x_N, y_N)\}$. We denote by $f: \mathbb{R}^d \xrightarrow{} [0,1]$ the fixed classifier for which recourse is to be determined. We denote the set of outcomes by $y(x)\in \{0,1\}$, where $y=1$ indicates the desirable outcome. Moreover, $\hat{y} = \mathbb{I}[f(x) > \theta]$ is the predicted class, where $\mathbb{I}[\cdot]$ denotes the indicator function and $\theta$ is a threshold (e.g., 0.5). Our goal is to find a set of actionable changes in order to improve the outcomes of instances $x$, which are assigned an undesirable prediction under $f$. Moreover, one typically defines a distance measure in inputs space $c: \mathbb{R}^d \times \mathbb{R}^d \xrightarrow{} \mathbb{R}_+$. We discuss typical choices for $c$ in Section \ref{section:benchmarking_process}.

Assuming inputs are pairwise statistically independent, the recourse problem is defined as follows:
\begin{align}
\begin{split}
    \delta_x^* &= \argmin_{\delta_x \in \mathcal{A}_d}  c\big(x, \CF \big)  \text{ s.t. } \CF = \F + \delta_x, f(x+\delta_x) > \theta,
\end{split}
\tag{$\mathbf{I}$}
\label{equation:problem_independent}
\end{align}
% \todo[author=Klaus]{Formally incorrect: (a) When writing $a = \argmin (\ldots)$, $a$ should not be part of $\ldots$ (you might do so, but you would end up in hell of an equation system and that's not what you intend. here). (b) The symbol below the $\argmin$ defines the output. So this equation states, that $\xC$ is assigned with the optimal $\delta_x$.
% {\bf Suggestion: }
% \begin{align*}
%     \delta_x^* &= \argmin_{\delta_x\in\mathcal{A}_d}  c\big(x, x^{F} + \delta_x \big) \\
%     \xC &= x + \delta_x^*
% \end{align*}}
where $\mathcal{A}_d$ is the set of admissible changes made to the factual input $x$. For example, $\mathcal{A}_d$ could specify that no changes to sensitive attributes such as \texttt{age} or \texttt{sex} may be made. 
% \todo[author=Klaus]{Question: is $\mathcal{A}_d$ independent of $x$? E.g. in a medical setting telling a 49kg weighting Person to lose 50kg is different from telling this a 220kg weighting one.}
% \todo[author=Martin]{Usually, we only look at global feature by feature
% constraints; i.e.\ weight $\in [40,250]$.}
For example, using the independent input assumption, existing approaches  \citep{ustun2019actionable} use mixed-integer linear programming to find counterfactual explanations.  In the next paragraph, we present a problem formulation that relaxes the strong independence assumption by introducing generative models.

\subsection{Recourse for Correlated Inputs} \label{section:background_recourse_correlated}
We assume the factual input $x \in \mathcal{X} = \mathbb{R}^d$ is generated by a generative model $g$ such that:
\begin{equation*}
    x = g(z),
\end{equation*}
% \todo[author=Klaus]{I do not get what the $D$ in $\delta_x^D$ is and why $D=1$ means independence. Can we define / clarify this?}
where $z \in \mathcal{Z} = \mathbb{R}^k$ are latent codes. We denote the counterfactual explanation in an input space by $\CF = x + \delta_x$. Thus, we have $\CF = x + \delta_x = g(z + \delta_z)$. Assuming inputs are dependent,
we can rewrite the recourse problem in \eqref{equation:problem_independent} to faithfully capture those dependencies using the generative model $g$:
\begin{align}
\begin{split}
    & \delta_z^* = \argmin_{\delta_z \in \mathcal{A}_k}  c\big(x, \CF \big)  \text{ s.t. } \CF = g(z + \delta_z), f(\CF) > \theta,
\end{split}
\tag{$\mathbf{D}$}
\label{equation:problem_dependent}
\end{align}
% \todo[author=Klaus]{I Have to bring up the formalism again: see \eqref{equation:problem_independent}.}
% \todo[author=Klaus, color=yellow]{(in my opinion, the math looks much better this way. Just one small thing:
% $\bar{x}$ is often used the mean. Consider $\tilde{x}$ or $\hat{x}$ (the latter one is often used for estimates) as an alternative to meet the readers expectation.}
where $\mathcal{A}_k$ is the set of admissible changes in the $k$-dimensional latent space. For example, $\mathcal{A}_k$ would ensure that the counterfactual latent space lies within range of $z$. The problem in \eqref{equation:problem_dependent} is an abstraction from how the problem is usually solved in practice: most existing approaches first train a type of autoencoder model (e.g., a VAE), and then use the model's trained decoder as a deterministic function $g$ to find counterfactual explanations \citep{joshi2019towards,pawelczyk2019,mahajan2019preserving,downs2020interpretable,antoran2020getting}. Our benchmarked explanation models roughly fit in one of these two categories. 

\begin{figure*}[!h]
    \centering
    \begin{subfigure}[b]{\textwidth}
         \centering
    \includegraphics[scale=0.36]{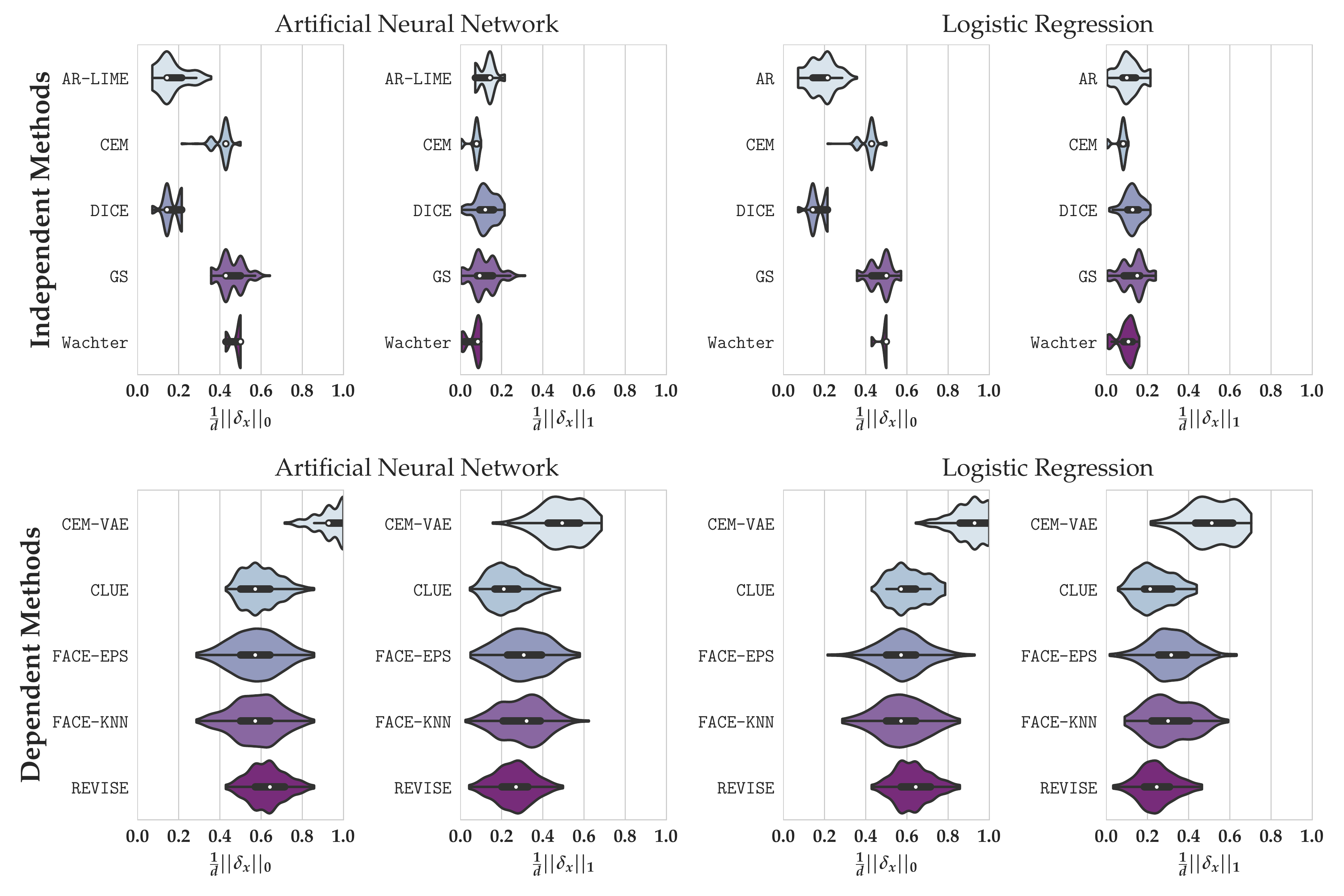}
    \caption{Adult Data}
    \vspace{-0.08cm}
    \label{fig:cost_comparison_adult}
\end{subfigure}
\begin{subfigure}[b]{\textwidth}
         \centering
    \includegraphics[scale=0.36]{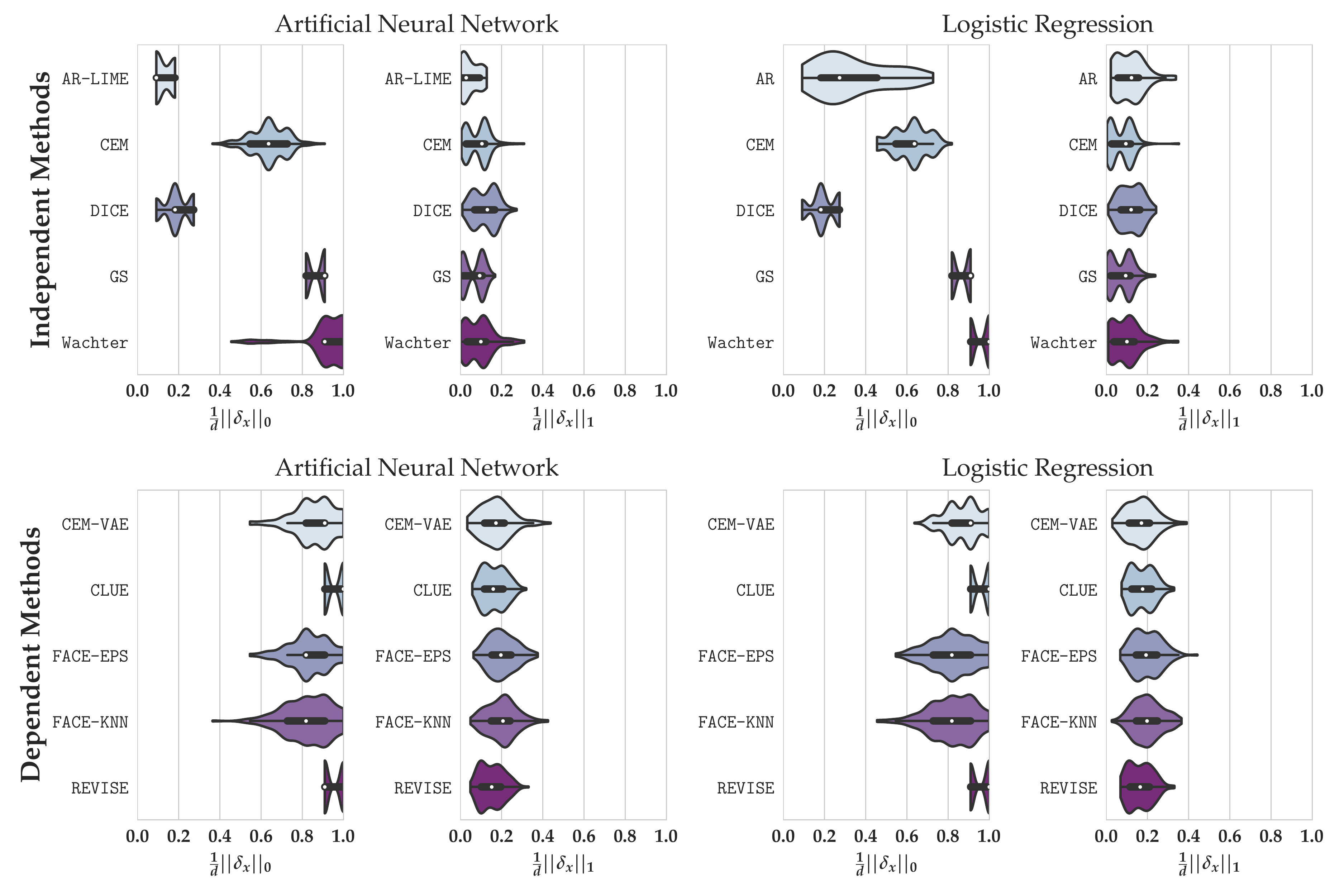}
    \caption{Give Me Some Credit Data}
    \label{fig:cost_comparison_gmc}
\end{subfigure}
\caption{Evaluating the distribution of costs of counterfactual explanations on 2 different data sets (the results on COMPAS are relegated to Appendix \ref{appendix:add_experiments}). For all instances with a negative prediction ($\{x \in \mathcal{D}: f(x) < \theta\}$), we plot the distribution of $\ell_0$ and $\ell_1$ costs of algorithmic recourse as defined in \eqref{eq:costs} for a logistic regression and an artificial neural network classifier. The white dots indicate the medians (lower is better), and the black boxes indicate the interquartile ranges. We distinguish between independence based and dependence based methods. The results are discussed in Section \ref{section:evaluation}}
\label{fig:cost_comparison_adult_gmc}
\end{figure*}

\section{Benchmarking Process} \label{section:benchmarking_process}
In this Section, we provide a brief explanation model overview and introduce a variety of explanation measures used to evaluate the quality of the generated counterfactual explanations. In Table \ref{tab:info_models} we present a concise explanation model overview.
\begin{table*}[bt]
    \centering
    %\begin{tabularx}{\textwidth}{cccccc}
    \resizebox{\columnwidth}{!}{%
    \begin{tabular}{ccccccc}
    \toprule
         Approach & Method &  Model Type & Algorithm & Immutable & Categorical & Other  \\
         \cmidrule(lr){1-1} \cmidrule(lr){2-2}  \cmidrule(lr){3-7}
         \multirow{6}{*}{Independent (\textbf{I})} & \texttt{AR} & Linear & Integer Prog.\ & Yes & Binary & Direction of change \\
         & \texttt{AR--LIME} & Agnostic & Integer Prog.\  & Yes & Binary & Direction of change \\
         & \texttt{CEM} & Gradient based & Gradient based & No & No & None \\
         & \texttt{DICE} & Gradient Based & Gradient based & Yes & Binary & Generative model \\
         & \texttt{GS} & Agnostic & Random search & Yes & Binary & None \\
         & \texttt{Wachter} & Gradient based & Gradient based & No & Binary & None \\
         \midrule
         \multirow{5}{*}{Dependent (\textbf{D}) } & \texttt{CEM-VAE} & Gradient based & Gradient based & No & No & Gen.\ Model regularizer \\
         & \texttt{CLUE} & Gradient based & Gradient based & No & No & Generative model \\
         & \texttt{FACE--EPS} & Agnostic & Graph search & Binary & Binary & CE is from data set \\
         & \texttt{FACE--KNN} & Agnositc & Graph search & Binary & Binary & CE is from data set \\
         & \texttt{REVISE} & Gradient based & Gradient based  & Binary & Binary & Generative model \\
    \bottomrule
    \end{tabular}
    }
    \caption{Explanation method summary: we categorize different approaches based on their underlying assumptions and list what kind of ML model they work with (Model Type), the Method's underlying algorithm (Algorithm), whether the method can handle immutable features (Immutable), whether it can handle categorical features (Categorical) and any other outstanding characteristics (Other).}
    \label{tab:info_models}
\end{table*}

\subsection{Counterfactual Explanation Methods}
\label{sec:expl_models}
\paragraph{\textbf{\texttt{AR} $\mathbf{(I)}$}}
\citet{Ustun2019ActionableRI} provide a method to generate minimal cost actions $\delta^*_x$ for linear classification models such as logistic regression models. \texttt{AR} requires the linear model's coefficients, and uses these coefficients for its search for counterfactual explanations. To provide reasonable actions it is possible to restrict $\delta_x^*$ to \emph{user--specified constraints} (e.g., \textit{has\_phd} can only change from \textit{False} to \textit{True}) or to set a subset of inputs as \emph{immutable} (e.g., \textit{age}). The problem to find these changes is a discrete optimization problem. Given a set of actions, \texttt{AR} finds the action which minimizes a defined cost function, using integer programming solvers like \texttt{CPLEX} or \texttt{CBC}.

\paragraph{\textbf{\texttt{AR--LIME} $\mathbf{(I)}$ }}
Most classification tasks do not have linearly separable classes and complex non--linear models usually provide more accurate predictions. Non--linear models are not per se interpretable and usually do not provide coefficients similar to linear models. We use a \textbf{reduction} to apply \texttt{AR} to non--linear models by computing a local linear approximation for the point of interest $x$, using \texttt{LIME} \cite{ribeiro2016should}. For an arbitrary black--box model $f$, \texttt{LIME} estimates post--hoc local explanations in form of a set of linear coefficients per instance. Using the coefficients we apply \texttt{AR}.

\paragraph{\textbf{\texttt{CEM} $\mathbf{(I)}$ }}
\citet{Dhurandhar2018} use an elastic--net regularization inspired objective to find low-cost counterfactual instances. Different weights can be assigned to $\ell_1$ and $\ell_2$ norms, respectively. There exists no immutable feature handling. However, we provide support for their VAE type regularizer, which should help ensure that counterfactual instances look more realistic.

\paragraph{\textbf{\texttt{CLUE}  $\mathbf{(D)}$ }}
\citet{antoran2020getting} propose \texttt{CLUE}, a generative recourse model that takes a classifier's uncertainty into account. This model suggests feasible counterfactual explanations that are likely to occur under the data distribution. The authors use a  variational autoencoder (VAE) to estimate the generative model. Using the VAE's decoder, \texttt{CLUE} uses an objective that guides the search of CEs towards instances that have low uncertainty measured in terms of the classifier's entropy.

\paragraph{\textbf{\texttt{DICE}  $\mathbf{(I)}$ }} \citet{Mothilal2020ExplainingML} suggest \texttt{DICE}, which is an explanation model that seeks to generate minimum costs counterfactual explanations according to \eqref{equation:problem_independent} subject to a diversity constraint which aims to promote a diverse set of counterfactual explanations. Diversity is achieved by using the whole range of suggested changes, while still keeping proximity to a given input. Regarding the optimization problem, \texttt{DICE} uses gradient descent to find a solution that trades-off proximity and diversity. Domain knowledge -- in form of feature ranges or immutability constraints -- can be added.

\paragraph{\textbf{\texttt{FACE}  $\mathbf{(D)}$ }}
The authors of \cite{Poyiadzi2020} provide \texttt{FACE}, which uses a shortest path algorithm (for graphs) to find counterfactual explanations from  high--density regions. Those explanations are actual data points from either the training or test set. Immutability constraints are enforced by removing incorrect neighbors from the graph. We implemented two variants of this model: the first variant uses an epsilon--graph (\texttt{FACE--EPS}), whereas the second variant uses a knn--graph (\texttt{FACE--KNN}).

\paragraph{\textbf{\texttt{Growing Spheres (GS)}  $\mathbf{(I)}$ }}
\texttt{Growing Spheres} -- suggested in \citep{laugel2017inverse} -- is a random search algorithm, which generates samples around the factual input point until a point with a corresponding counterfactual class label was found. The random samples are generated around $x$ using growing hyperspheres. For binary input dimensions, the method makes use of Bernoulli sampling. Immutable features are readily specified by excluding them from the search procedure.

\paragraph{\textbf{\texttt{REVISE} $\mathbf{(D)}$}} \citet{joshi2019towards} propose a generative recourse model. This model suggests feasible counterfactual explanations that are likely to occur under the data distribution. The authors use a  variational autoencoder (VAE) to estimate the generative model. Using the VAE's decoder, \texttt{REVISE} uses the latent space to search for CEs. No handling of immutable features exists.

\paragraph{\textbf{\texttt{Wachter et al.\ (Wachter)} $\mathbf{(I)}$}}
The optimization approach suggested by \citet{Wachter2017CounterfactualEW} generates counterfactual explanations by minimizing an objective function using gradient descent to find counterfactuals $\CF$ which are as close as possible to $x$. Closeness is measured in $\ell_1$-norm.

% The objective is stated in the following Equation (\ref{equation:wachter}).
% \begin{align}
% \begin{split}
%     & \arg \min_{\CF} \max_{\lambda} \lambda(f_w(\CF)-y')^2+d(x_i,\CF)
% \end{split}
% \tag{$\mathbf{W}$}
% \label{equation:wachter}
% \end{align}
% Where $d(\cdot, \cdot)$ measures the distance between factual $x$ and counterfactual $\CF$, and $f_w(\cdot)$ is a loss function with fixed weights $w$. The maximization over $\lambda$ is done iteratively by minimizing $\CF$ and increasing $\lambda$ till we reach target class $y'$.

\begin{table*}[tb]
\centering
\begin{subtable}[b]{\textwidth}

\centering
\resizebox{\columnwidth}{!}{%
\begin{tabular}{ccrrrrrrrrrr}
\toprule
\multicolumn{2}{c}{} & \multicolumn{5}{c}{Artificial Neural Network} & \multicolumn{5}{c}{Logistic Regression} \\
\cmidrule(lr){3-7} \cmidrule(lr){8-12}
Data Set  & Method &   \textit{yNN} &  redund. &  violation &  success &  $\overline{t}(s)$ &   \textit{yNN} &  redund. &  violation &  success &  $\overline{t}(s)$ \\
\cmidrule(lr){1-1} \cmidrule(lr){2-2} \cmidrule(lr){3-7} \cmidrule(lr){8-12}

\multirow{5}{*}{Adult} & \texttt{AR(--LIME)} &          0.62 &     \textbf{0.00} &       0.14 &     0.28 &               1.59 &          \textbf{0.72} &     0.67 &       0.13 &     0.52 &              10.49 \\
    & \texttt{CEM} &          0.26 &     3.96 &       0.66 &     \textbf{1.00} &               1.10 &          0.20 &     3.98 &       0.66 &    \textbf{1.00} &               0.92 \\
    & \texttt{DICE} &          \textbf{0.71} &     0.53 &       0.17 &     \textbf{1.00} &               0.13 &          0.58 &     \textbf{0.51} &       0.23 &     \textbf{1.00} &               0.13 \\
    & \texttt{GS} &          0.30 &     3.77 &       \textbf{0.09} &     \textbf{1.00} &               \textbf{0.01} &          0.30 &     3.94 &       \textbf{0.10} &     \textbf{1.00} &               \textbf{0.01} \\
    & \texttt{Wachter} &          0.23 &     4.45 &       0.83 &     0.50 &              15.72 &          0.16 &     1.67 &       0.94 &     \textbf{1.00} &               0.03 \\
\midrule
\multirow{5}{*}{GMC} & \texttt{AR(--LIME)} &          0.89 &     \textbf{0.00} &       0.29 &     0.07 &               0.55 &         \textbf{1.00} &     2.33 &       \textbf{0.14} &     0.39 &               3.42 \\
    & \texttt{CEM} &          \textbf{0.95} &     5.46 &       0.65 &     \textbf{1.00} &               0.97 &          0.74 &     5.07 &       0.67 &     \textbf{1.00} &               0.87 \\
    & \texttt{DICE} &          0.90 &     0.58 &       0.27 &     \textbf{1.00} &               0.28 &          0.88 &     \textbf{0.61} &       0.27 &     \textbf{1.00} &               0.29 \\
    & \texttt{GS} &          0.40 &     6.64 &       \textbf{0.17} &     \textbf{1.00} &               \textbf{0.01} &          0.49 &     5.29 &       0.17 &     \textbf{1.00} &               \textbf{0.01} \\
    & \texttt{Wachter} &          0.58 &     6.56 &       0.71 &     \textbf{1.00} &               0.02 &          0.59 &     6.12 &       0.83 &     \textbf{1.00} &               \textbf{0.01} \\
\bottomrule

\end{tabular}
}
\caption{Independence based methods}
\label{tab:method_comparison_independent}
\end{subtable}
\vspace{0.05cm}

\vfill

\begin{subtable}[b]{\textwidth}

\centering
\resizebox{\columnwidth}{!}{%
\begin{tabular}{ccrrrrrrrrrr}
\toprule
\multicolumn{2}{c}{} & \multicolumn{5}{c}{Artificial Neural Network} & \multicolumn{5}{c}{Logistic Regression} \\
\cmidrule(lr){3-7} \cmidrule(lr){8-12}
Data Set  & Method &  \textit{yNN} &  redund. &  violation &  success &  $\overline{t}(s)$ &  \textit{yNN} &  redund. &  violation &  success &  $\overline{t}(s)$ \\
\cmidrule(lr){1-1} \cmidrule(lr){2-2} \cmidrule(lr){3-7} \cmidrule(lr){8-12}

\multirow{5}{*}{Adult} & \texttt{CEM--VAE} &          0.12 &     9.68 &       1.82 &     \textbf{1.00} &               \textbf{0.93} &          0.43 &    10.05 &       1.80 &     \textbf{1.00} &               \textbf{0.81} \\
    & \texttt{CLUE} &          \textbf{0.82} &     8.05 &       \textbf{1.28} &     \textbf{1.00}&               2.70 &          0.33 &     7.30 &       1.33 &     \textbf{1.00}&               2.56 \\
    & \texttt{FACE--EPS} &          0.65 &     5.19 &       1.45 &     0.99 &               4.36 &          \textbf{0.64} &     5.11 &       1.44 &     0.94 &               4.35 \\
    & \texttt{FACE--KNN} &          0.60 &     \textbf{5.11} &       1.41 &     \textbf{1.00} &               4.31 &          0.57 &     \textbf{4.97} &       1.38 &     1.00 &               4.31 \\
    & \texttt{REVISE} &          0.20 &     8.65 &       1.33 &     1.00 &               8.33 &          0.62 &     7.92 &       \textbf{1.23} &     \textbf{1.00} &               7.52 \\
\midrule
\multirow{5}{*}{GMC} & \texttt{CEM--VAE} &          \textbf{1.00} &     8.40 &       \textbf{0.66} &     \textbf{1.00} &               \textbf{0.87} &          \textbf{1.00} &     8.54 &       \textbf{0.36} &    \textbf{1.00} &               \textbf{0.88} \\
    & \texttt{CLUE} &          \textbf{1.00} &     9.39 &       0.90 &     0.93 &               1.91 &          \textbf{1.00} &     9.56 &       0.96 &     \textbf{1.00 }&               1.76 \\
    & \texttt{FACE--EPS} &          0.99 &     \textbf{8.06} &       0.99 &     \textbf{1.00} &              19.44 &          0.98 &     7.98 &       0.96 &    \textbf{1.00} &              19.50 \\
    & \texttt{FACE--KNN} &          0.98 &     9.00 &       0.98 &    \textbf{1.00} &              15.87 &          0.98 &     \textbf{7.88} &       0.95 &     \textbf{1.00} &              16.09 \\
    & \texttt{REVISE} &          \textbf{1.00} &     9.50 &       0.97 &     \textbf{1.00} &               4.56 &          \textbf{1.00} &     9.59 &       0.96 &     1.00 &               3.76 \\
\bottomrule

\end{tabular}
}
\caption{Dependence based methods}
\label{tab:method_comparison_dependent}

\end{subtable}

\caption{Summary of a subset of results for independence and dependence based methods. For all instances with a negative prediction ($\{x \in \mathcal{D}: f(x) < \theta\}$), we compute counterfactual explanations for which we then measure $\text{yNN}$ (higher is better), redundancy (lower is better), violation (lower is better), success rate (higher is better) and time (lower is better). We distinguish between a logistic regression and an artificial neural network classifier. Detailed descriptions of these measures can be found in Section \ref{section:benchmarking_process}. The results are discussed in Section \ref{section:evaluation}.}

\label{table:all_results}
\end{table*}
\subsection{Evaluation Measures for Counterfactual Explanation Methods} \label{section:measures}
%\todo[author=Tobi]{@Martin remove first sentence? Feels unessary / assumptious}
%Comparisons between different CE methods are complex because algorithms optimize their own objective functions.
As algorithmic recourse is a multi--modal problem we introduce a variety of measures to evaluate the methods' performances. We use six baseline evaluation measures. Besides distance measures it is important to consider measures that emphasize the \emph{quality} of recourse.

%We want to evaluate how many of the proposed feature changes were necessary, and how often the specified user constraints were violated. Taking the attention more to computational performance, to measure real-life practicability, we will check the average computation time to generate a counterfactual example. Additionally, we will track the proportion of generated examples which do not change the classification of the given black-box model, since some CE methods' provided implementations do not guarantee that every generated example alter the prediction as requested. We will introduce a metric which explores the neighbourhood of a counterfactual example and evaluate how many of its neighbours are labeled with the same class.

\paragraph{\textbf{Costs}}
When answering the question of generating the nearest counterfactual explanation, it is essential to define the distance of the factual $\F$ to the nearest counterfactual $\CF$. The literature has formed a consensus to use either the normalized $\ell_0$ or $\ell_1$ norm or any convex combination thereof (see for example \cite{rawal2020individualized,mothilal2020fat,pmlr-v124-pawelczyk20a,karimi2019model,ustun2019actionable,wachter2017counterfactual}). The $\ell_0$ norm puts a restriction on the number of feature changes between factual and counterfactual instance, while the $\ell_1$ norm restricts the average change:
\begin{align}
c_0(\CF, \F) &= \frac{1}{d} \lVert x - \CF \rVert_0 = \frac{1}{d} \lVert \delta_x \rVert_0, & 
c_1(\CF, \F) &= \frac{1}{d} \lVert x - \CF \rVert_1 = \frac{1}{d} \lVert  \delta_x \rVert_1. 
\label{eq:costs}
\end{align}

\paragraph{\textbf{Constraint violation}}
This measure counts the number of times the CE method violates user-defined constraints. Depending on the data set, we fixed a list of features which should not be changed by the used method (e.g., \emph{sex}, \emph{age} or \emph{race}).

\paragraph{\textbf{yNN}} We use a measure that evaluates how much data support CEs have from positively classified instances. Ideally, CEs should be close to positively classified individuals which is a desideratum formulated by \citet{laugel2019dangers}. We define the set of individuals who received an undesirable prediction under $f$ as $H^- := \{x \in \mathcal{D}:f(x) < \theta \}$. The counterfactual instances (instances for which the label was successfully changed) corresponding to the set $H^-$ are denoted by $\check{H}^-$. We use a measure that captures how differently neighborhood points around a counterfactual instance $\CF$ are classified:
\begin{equation}
\text{yNN} = 1 - \frac{1}{nk} \sum_{i \in \check{H}^-}^n \sum_{j \in \text{kNN}(\CF_i)} \big|f_b(\CF_i) - f_b(x_{j})\big|,  \label{eq:measure2}
\end{equation}
where kNN denotes the $k$-nearest neighbours of $x$, and $f_b(x) = \mathbb{I}[f(x) > 0.5]$ is the binarized classifier. Values of \text{yNN} close to 1 imply that the neighbourhoods around the counterfactual explanations consists of points with the same predicted label, indicating that the neighborhoods around these points have already been reached by positively classified instances. We use a value of $k:=5$, which ensures sufficient data support from the positive class.

\paragraph{\textbf{Redundancy}}
We evaluate how many of the proposed feature changes were not necessary. This is a particularly important criterion for independence--based methods. We measure this by \emph{successively} flipping one value of $\CF$ after another back to $\F$, and then we inspect whether the label flipped from $1$ back to $0$: e.g., we check whether flipping the value for the second dimension would change the counterfactual outcome $1$ back to the predicted factual outcome of $0$: $\mathbb{I}[f_b \big([\CF_1, x_2, \CF_3, \dots, \CF_d]\big)=0]$. If the predicted outcome does not change, we increase the redundancy counter, concluding that a sparser counterfactual explanation could have been found. We iterate this process over all dimensions of the input vector.\footnote{We do not consider all possible subsets of changes.}  A low number indicates few redundancies across counterfactual instances.

\paragraph{\textbf{Success Rate}} Some generated counterfactual explanations do not alter the predicted label of the instance as anticipated. To keep track how often the generated CE does hold its promise, the success rate shows the fraction of respective models' correctly determined counterfactuals.

\paragraph{\textbf{Average Time}} By measuring the average time a CE method needs to generate its result, we evaluate the effectiveness and feasibility for real--time prediction settings.

\section{Experimental Evaluation}\label{section:evaluation}

Using \texttt{CARLA} we conduct extensive empirical evaluations to benchmark the presented counterfactual explanations methods using three real-world data sets. Our main findings are displayed in Figure \ref{fig:cost_comparison_adult_gmc}, and Table \ref{table:all_results}. We split the benchmarking evaluation by CE method category. In the following Sections, we provide an overview over the used data sets (see Table \ref{tab:info_data_set}) and the classification models. Detailed information on hyperparameter search for the CE methods is provided in Appendix \ref{appendix:hyperparameters}.

\paragraph{Data sets}
The \textbf{Adult} data set \cite{Dua:2019} originates from the 1994 Census database, consisting of 14 attributes and 48,842 instances. The classification consists of deciding whether an individual has an income greater than 50,000 USD/year. Since several CE methods cannot handle non-binary categorical data, we binarized these features by partitioning them into the most frequent value, and its counterpart (e.g., \emph{US} and \emph{Non-US}, \emph{Husband} and \emph{Non-Husband}).
The features \textit{age}, \textit{sex} and \textit{race} are set as immutable.
The \textbf{Give Me Some Credit} (GMC) data set \cite{Kaggle2011} from a 2011 Kaggle Competition is a credit scoring data set, consisting of 150,000 observations and 11 features. The classification task consists of deciding whether an instance will experience financial distress within the next two years (\textit{SeriousDlqin2yrs} is 1) or not. We dropped missing data, and set \textit{age} as immutable.
\begin{table*}[!htb]
    \centering
    \resizebox{\columnwidth}{!}{%
    \begin{tabular}{cccccc}
    \toprule
         Data Set &  Task & Positive Class & Size $(N \; | \; d)$ & Features & Immutable Features  \\
         \cmidrule(lr){1-1}  \cmidrule(lr){2-6}
         Adult & Predict Income & High Income (24\%) & (45,222 | 20) & Work, Education, Income & Sex, Age, Race \\
         %\cmidrule(lr){1-6}
         COMPAS & Predict Recidivism & No Recid.\ (65\%) & (10,000 | 8) & Crim.\ History, Jail \& Prison Time &  Sex, Race \\
         %\cmidrule(lr){1-6}
         GMC & Predict Financial Distress  & No deficiency (93\%) & (150,000 | 11) & Pay.\ History, Balance, Loans & Age \\
    \bottomrule
    \end{tabular}
    }
    \caption{Summarizing the used data sets, where $N$ and $d$ are the number of samples and input dimension after processing of the data. Results on the COMPAS data set are relegated to Appendix \ref{appendix:add_experiments}.}
    \label{tab:info_data_set}
\end{table*}

\paragraph{Black-box models}
We briefly describe how the black--box classifiers $f$ were trained. \texttt{CARLA} supports different ML libraries (e.g., Pytorch, Tensorflow) to estimate these classifiers as the implementations of the various explanation methods work particular ML libraries only. 
The first model is a multi-layer perceptron, consisting of three hidden layers with 18, 9 and 3 neurons, respectively. To allow a more extensive comparison (\texttt{AR} only works on linear models) between CE methods, we chose logistic regression models as the second classification model for which we evaluate the CE methods. Detailed information on the classifiers' training for each data set is provided in Appendix \ref{appendix:ml_classifiers}.

\paragraph{Benchmarking} For \textbf{independence based} methods, we find that no one single CE method outperformed all its competitors. This is not too surprising since algorithmic recourse is a multi--modal problem. Instead, we found that some methods dominated certain measures across all data sets. \texttt{AR}, \texttt{AR--LIME}, \texttt{DICE} performed strongest with respect to $\ell_0$ (see the top left panels in Figures \ref{fig:cost_comparison_adult} and \ref{fig:cost_comparison_gmc}). \texttt{AR--LIME} does so despite our use of the \texttt{LIME} reduction. Therefore, it makes sense that \texttt{AR}, \texttt{AR-LIME} and \texttt{DICE} offer the lowest \emph{redundancy} scores (Table \ref{tab:method_comparison_independent}). \texttt{CEM} performed strongest with respect to the overall cost measure $\ell_1$ across data sets. \texttt{GS} is the clear winner when it comes to the measurement of time (Table \ref{tab:method_comparison_independent}). Since the algorithm behind \texttt{GS} is based on a rather rudimentary sampling strategy, we expect that savvier sampling strategies should boost its cost performance significantly. 

%\todo[]{@Martin: The above 0.65 ynn measure for dependence based is not really true for both FACEs. Maybe highlight the low yNN Measures for CEM and REVISE on Adult ANN instead?}
For \textbf{dependence based} methods, the results are mixed as well. While \texttt{CLUE} and \texttt{REVISE} are the winner with respect to the cost of recourse ($\ell_1$), the margins between these generative recourse models and the graph-based ones (\texttt{FACE}) are small (Figure \ref{fig:cost_comparison_adult_gmc}). The \texttt{FACE-EPS} method performs strongest with respect to the $ynn$ measure (usually well above 0.60) (Table \ref{tab:method_comparison_dependent}) indicating that the generated CEs have sufficient data support from positively classified individuals relatively to the remaining dependence--based methods. As expected, the ynn measures are on average higher for the dependence based methods. This suggest that dependence based CEs are less often outliers. Notably, \texttt{CLUE} and \texttt{REVISE} perform best with respect to $\ell_1$ (with \texttt{REVISE} being the clear winner on 3 out of 4 cases), while they perform worst on $\ell_0$ -- likely due to the decoder's imprecise reconstruction. In this respect, it is not surprising that these methods have average redundancy values that are up to twice as high as those by \texttt{FACE}. Finally, the generative model approaches (\texttt{CEM-VAE}, \texttt{CLUE}, \texttt{REVISE}) performed best with respect to time since the autoencoder training time amortizes with more samples.

\section{Conclusion and Broader Impact of \texttt{CARLA}}\label{section:conclusion}
The current implementations of recourse methods, mentioned in Section \ref{sec:expl_models} are based on the original implementation of the respective research groups. Researchers mostly implement their experiments and models for specific ML frameworks and data sets. For example, some explanation methods are restricted to Tensorflow and are not applicable to Pytorch models. In the future, we will extend \texttt{CARLA} to decouple each recourse method from the frameworks and data contraints.

When trying to combine different CE methods into a common benchmarking framework we encountered the following issues: First, a great number of repositories only contain remarks about installation and script calls to recreate the results from the corresponding research papers. Second, missing information about interfaces for data sets or black--box models further complicated the process of integrating different CE methods into the benchmarking workflow. In order to add more CE methods and data sets to \texttt{CARLA}, we are currently in contact with several authors in this exciting and rapidly growing field. With a growing open-source community, \texttt{CARLA} can evolve to be the main library for generating counterfactual explanations and benchmarks for recourse methods. Therefore we are continuously expanding the catalog of explanation methods and data sets, and welcome researchers to add their own recourse methods to the library. To facilitate this process, we provide a step-by-step user-guide to integrate new CE methods into \texttt{CARLA}, which we present in Appendix \ref{appendix:carla_software}. 

The rapidly growing number of available CE methods calls for standardized and efficient ways to assure the quality of a new technique in comparison with other approaches on different data sets. Quality assurance is a key aspect of actionable recourse, since complex models and CE mechanisms can have a considerable impact on personal lives. In this work, we presented \texttt{CARLA}, a versatile benchmarking platform for the standardized and transparent comparison of CE methods on different integrated data sets. In the explainability field, \texttt{CARLA} bears the potential to help researchers and practitioners alike to efficiently derive more realistic and use--case--driven recourse strategies and assure their quality through extensive comparative evaluations. We hope that this work contributes to further advances in explainability research. %and AI in general.

%\section{Broader Impact Statement} \label{section:broader_impact}

\newpage

\bibliographystyle{ACM-Reference-Format}
\bibliography{main.bib}

\newpage

\section*{Checklist}

%%% BEGIN INSTRUCTIONS %%%
The checklist follows the references.  Please
read the checklist guidelines carefully for information on how to answer these
questions.  For each question, change the default \answerTODO{} to \answerYes{},
\answerNo{}, or \answerNA{}.  You are strongly encouraged to include a {\bf
justification to your answer}, either by referencing the appropriate section of
your paper or providing a brief inline description.  For example:
\begin{itemize}
  \item Did you include the license to the code and datasets? \answerYes{See Section~\ref{gen_inst}.}
  \item Did you include the license to the code and datasets? \answerNo{The code and the data are proprietary.}
  \item Did you include the license to the code and datasets? \answerNA{}
\end{itemize}
Please do not modify the questions and only use the provided macros for your
answers.  Note that the Checklist section does not count towards the page
limit.  In your paper, please delete this instructions block and only keep the
Checklist section heading above along with the questions/answers below.
%%% END INSTRUCTIONS %%%

\begin{enumerate}

\item For all authors...
\begin{enumerate}
  \item Do the main claims made in the abstract and introduction accurately reflect the paper's contributions and scope?
    %\answerTODO{}
    \answerYes{}. As we state in the abstract, our goal is to provide a Python framework for benchmarking counterfactual explanation methods. Users can easily evaluate our results by accessing our \href{https://github.com/indyfree/CARLA}{Github repository}, where we host our Python framework and our benchmarking results.
  \item Did you describe the limitations of your work?
    %\answerTODO{}
    \answerYes{}. In Section \ref{section:conclusion}, we discuss the current limitations of our approach. The counterfactual explanation methods are based on the original implementation of the respective research groups. Researchers mostly implement their experiments and models for specific ML frameworks and data sets. For example, some explanation methods are restricted to Tensorflow and are not applicable to Pytorch models.
  \item Did you discuss any potential negative societal impacts of your work?
    %\answerTODO{}
    \answerNA{}. We discuss the broader impact of our benchmarking library in Section \ref{section:conclusion}; we mainly see positive impacts on the literature of algorithmic recourse.
  \item Have you read the ethics review guidelines and ensured that your paper conforms to them?
    %\answerTODO{}
    \answerYes{}. We have read the ethics review guidelines and attest that our paper conforms to the guidelines.
\end{enumerate}

\item If you are including theoretical results...
\begin{enumerate}
  \item Did you state the full set of assumptions of all theoretical results?
    %\answerTODO{}
    \answerNA{}. We did not provide theoretical results.
	\item Did you include complete proofs of all theoretical results?
    %\answerTODO{}
    \answerNA{}. We did not provide theoretical results.
\end{enumerate}

\item If you ran experiments...
\begin{enumerate}
  \item Did you include the code, data, and instructions needed to reproduce the main experimental results (either in the supplemental material or as a URL)?
    %\answerTODO{}
    \answerYes{}. Details of implementations, data sets and instructions can be found here: Appendices \ref{appendix:carla_software}, \ref{appendix:ml_classifiers}, \ref{appendix:hyperparameters}, and our \href{https://github.com/indyfree/CARLA}{Github repository}.
  \item Did you specify all the training details (e.g., data splits, hyperparameters, how they were chosen)?
    %\answerTODO{}
    \answerYes{}. Please see Appendices \ref{appendix:hyperparameters} and \ref{appendix:ml_classifiers}.
	\item Did you report error bars (e.g., with respect to the random seed after running experiments multiple times)?
    %\answerTODO{}
    \answerYes{}. Error bars have been reported for our cost comparisons in terms of the 25th and 75ht percentiles of the cost distribution, see for example Figure \ref{fig:cost_comparison_adult_gmc}.
	\item Did you include the total amount of compute and the type of resources used (e.g., type of GPUs, internal cluster, or cloud provider)?
    %\answerTODO{}
    \answerYes{}. All models are evaluated on an i7-8550U CPU with 16 Gb RAM, running on Windows 10.
\end{enumerate}

\item If you are using existing assets (e.g., code, data, models) or curating/releasing new assets...
\begin{enumerate}
  \item If your work uses existing assets, did you cite the creators?
   % \answerTODO{}
    \answerYes{}. The data sets, which are publicly available are appropriately cited in Section~\ref{section:evaluation}. We cite and link to any additional code used, for example \cite{antoran2020getting}.
  \item Did you mention the license of the assets?
    %\answerTODO{}
    \answerYes{}. All assets are publicly available and attributed.
  \item Did you include any new assets either in the supplemental material or as a URL?
    %\answerTODO{}
    \answerYes{}. Our implementation and code is accessible through our \href{https://github.com/indyfree/CARLA}{Github repository}.
  \item Did you discuss whether and how consent was obtained from people whose data you're using/curating?
    %\answerTODO{}
    \answerNA{}. We use publicly available data sets without any personal identifying information.
  \item Did you discuss whether the data you are using/curating contains personally identifiable information or offensive content?
    %\answerTODO{}
    \answerNA{}. We use publicly available data sets without any personal identifying information.
\end{enumerate}

\item If you used crowdsourcing or conducted research with human subjects...
\begin{enumerate}
  \item Did you include the full text of instructions given to participants and screenshots, if applicable?
    %\answerTODO{}
    \answerNA{}. We did not use crowdsourcing or conduct research with human subjects. 
  \item Did you describe any potential participant risks, with links to Institutional Review Board (IRB) approvals, if applicable?
    %\answerTODO{}
    \answerNA{}. We did not use crowdsourcing or conduct research with human subjects. 
  \item Did you include the estimated hourly wage paid to participants and the total amount spent on participant compensation?
    %\answerTODO{}
    \answerNA{}. We did not use crowdsourcing or conduct research with human subjects. 
\end{enumerate}
\end{enumerate}

\newpage

% Just as a default place to write about the software part. Can replaced everywhere
\appendix
\section{\texttt{CARLA}'s Software Interface}\label{appendix:carla_software}
In the following, we introduce our open-source benchmarking software \texttt{CARLA}. we describe the architecture in more detail and provide examples of different use-cases and their implementation.

\subsection{\texttt{CARLA}'s High Level Software Architecture}
The purpose of this Python library is to provide a simple and standardized framework to allow users to apply different state-of-the-art recourse methods to arbitrary data sets and black-box-models. It is possible to compare different approaches and save the evaluation results, as described in Section \ref{section:measures}. For research groups, \texttt{CARLA} provides an implementation interface to integrate new recourse methods in an easy-to-use way, which allows to compare their method to already existing methods.

\begin{figure*}[h]
    \centering
    \includegraphics[scale=0.32]{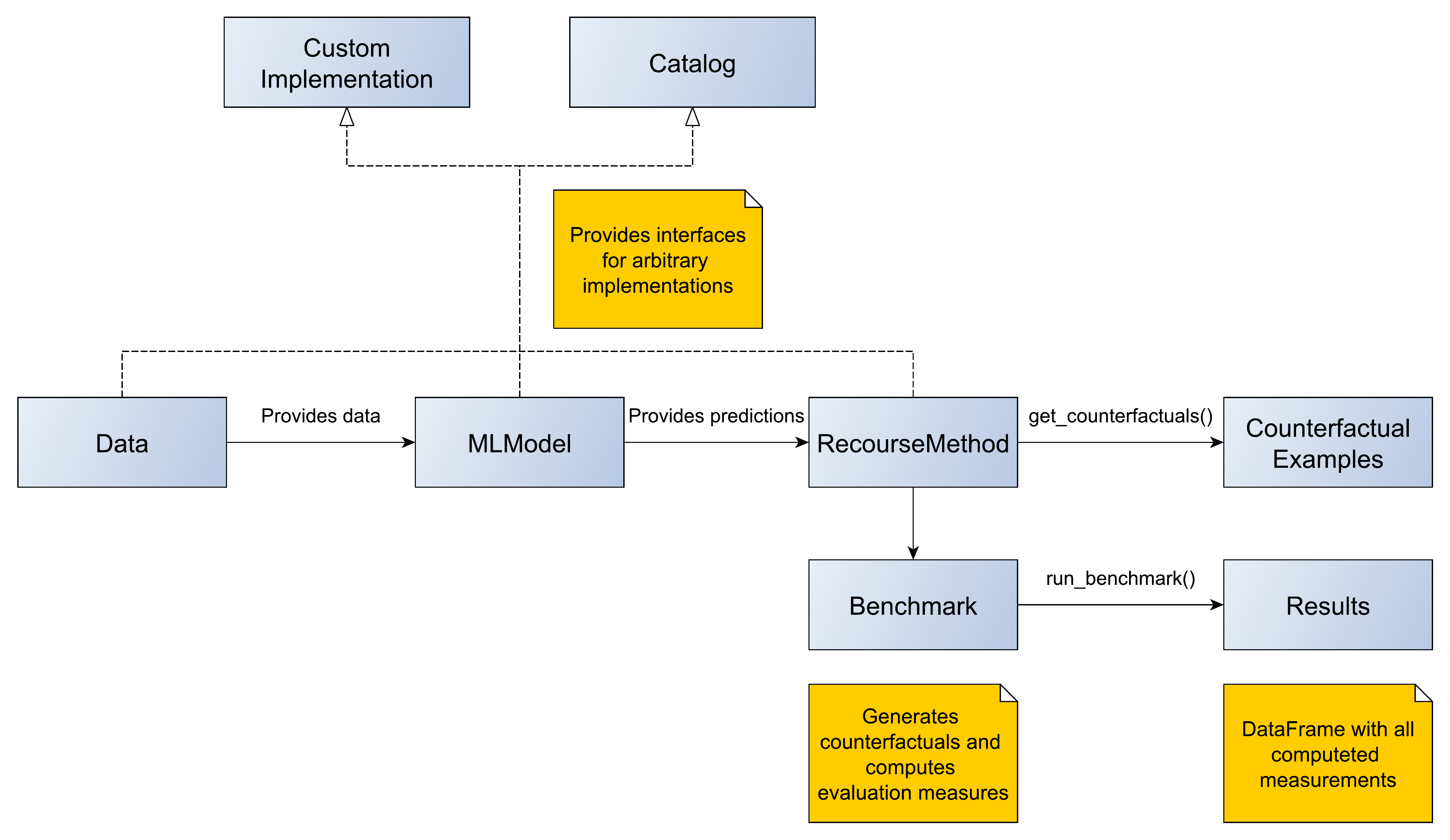}
    \caption{Architecture of the \texttt{CARLA} python library. The silver boxes show the individual objects that will be created to generate counterfactual explanations and evaluate recourse methods. Useful explanations to specific processes are illustrated as yellow notes. The dashed arrows are showing the different implementation possibilities, either use pre-defined catalog objects or provide custom implementation. All dependencies between these objects are visualised by solid arrows with an additional description.}
    \label{fig:carla_architecture}
\end{figure*}

A simplified visualization of the \texttt{CARLA} software architecture is depicted in Figure \ref{fig:carla_architecture}. For every component (\emph{Data}, \emph{MLModel}, and \emph{RecourseMethod}) the library provides the possibility to use existing methods from our catalog, or extend the users custom methods and implementations. The components represent an interface to the key parts in the process of generating counterfactual explanations. \textit{Data} provides a common way to access the data across the software and maintains information about the features. \textit{MLModel} wraps each black-box model and stores details on the encoding, scaling and feature order specific to the model. The primary purpose of \textit{RecourseMethod} is to provide a common interface to easily generate counterfactual examples.

Besides the possibility to use pretrained black-box-models and preprocessed data, \texttt{CARLA} provides an easy way to load and define own data sets and model structures independent of their framework (e.g., Pytorch, Tensorflow, sklearn). The following sections will give an overview and provide example implementations of different use cases.

\subsection{\texttt{CARLA} for Research Groups}
One of the most exciting features of \texttt{CARLA} is, that research groups can make use of the \textit{RecourseMethod}-wrapper to implement their own method to generate counterfactual examples. This opens up a way of standardized and consistent comparisons between different recourse methods. Strong and weak points of new algorithms can be stated, benchmarked and analysed in forthcoming publications with the help of \texttt{CARLA}.

In Figure \ref{fig:carla_example_recourse}, we show how an implementation of a custom recourse method can be structured. After defining the recourse method in the shown way, it can be used with the library to generate counterfactuals for a given data set and benchmark its results against other methods. Research groups have the choice to do this using our provided catalog of data sets, recourse methods and black-box models (Figure \ref{fig:carla_example1}) or use their own models and data sets (see Figures \ref{fig:carla_example_data} and \ref{fig:carla_example_model}).

\begin{figure*}[h]
    \centering
    \begin{lstlisting}[language=Python]
    
    from carla import RecourseMethod

    # Custom recourse implementations need to
    # inherit from the RecourseMethod interface
    class MyRecourseMethod(RecourseMethod):
        def __init__(self, mlmodel):
            super().__init__(mlmodel)
    	
        # Generate and return encoded and
        # scaled counterfactual examples
        def get_counterfactuals(self, factuals: pd.DataFrame):
    		[...]
    		return counterfactual_examples
    \end{lstlisting}
    \caption{Pseudo-implementation of the \texttt{CARLA} recourse method wrapper}
    \label{fig:carla_example_recourse}
\end{figure*}

\subsection{\texttt{CARLA} as a Recourse Library}
A common usage of the package is to generate counterfactual examples. This can be done by loading black-box-models and data sets from our provided catalogs, or by user-defined models and datasets via integration with the defined interfaces. Figure \ref{fig:carla_example1} shows an implementation example of a simple use-case, applying a recourse method to a pre-defined data set and model from our catalog. After importing both catalogs, the only necessary step is to describe the data set name (e.g., adult, give me some credit, or compas) and the model type (e.g., ann, or linear) the user wants to load. Every recourse method contains the same properties to generate counterfactual examples.

\begin{figure*}[h]
    \centering
    \begin{lstlisting}[language=Python]
    
    from carla import DataCatalog, MLModelCatalog
    from carla.recourse_methods import GrowingSpheres
    
    # 1. Load data set from the DataCatalog
    data_name = "adult"
    dataset = DataCatalog(data_name)
    
    # 2. Load pre-trained black-box model from the MLModelCatalog
    model = MLModelCatalog(dataset, "ann")
        
    # 3. Load recourse model with model specific hyperparameters
    gs = GrowingSpheres(model)
    
    # 4. Generate counterfactual examples
    factuals = dataset.raw.sample(10)
    counterfactuals = gs.get_counterfactuals(factuals)
    \end{lstlisting}
    \caption{Example implementation of \texttt{CARLA}, using the data and model catalog.}
    \label{fig:carla_example1}
\end{figure*}

To give users the possiblity to explore their own black-box-model on a custom data set, we implemented in \texttt{CARLA} easy-to-use interfaces, that are able to wrap every possible model or data set. These interfaces specify particular properties users have to implement, to be able to work with the library. Figure \ref{fig:carla_example_data} shows an example implementation of the data wrapper, and Figure \ref{fig:carla_example_model} depicts the same for an arbitrary black-box-model. After defining data set and black-box model classes, users simply need to call the canonical methods and generate counterfactual examples, similar to the process in Figure \ref{fig:carla_example1}.

\begin{figure*}[h]
    \centering
    \begin{lstlisting}[language=Python]
    
    from carla import Data
    from carla.recourse_methods import GrowingSpheres
    
    # Custom data set implementations need to inherit from the Data interface
    class MyOwnDataSet(Data):
        def __init__(self):
            # The data set can e.g. be loaded in the constructor
            self._dataset = load_dataset_from_disk()

        # List of all categorical features
        def categoricals(self):
            return [...]
    
        # List of all continous features
        def continous(self):
            return [...]
    
        # List of all immutable features which
        # should not be changed by the recourse method
        def immutables(self):
            return [...]
    
        # Feature name of the target column
        def target(self):
            return "label"
    
        # Non-encoded and  non-normalized, raw data set
        def raw(self):
            return self._dataset
    \end{lstlisting}
    \caption{Pseudo-implementation of the \texttt{CARLA} data wrapper}
    \label{fig:carla_example_data}
\end{figure*}

\begin{figure*}[h]
    \centering
    \begin{lstlisting}[language=Python]
    
    from carla import MLModel
    
    # Custom black-box models need to inherit from 
    # the MLModel interface
    class MyOwnModel(MLModel):
        def __init__(self, data):
            super().__init__(data)
            # The constructor can be used to load or build an 
            # arbitrary black-box-model
            self._mymodel = load_model()
    
            # Define a fitted sklearn scaler to normalize input data
            self.scaler = MySklearnScaler().fit()
    
            # Define a fitted sklearn encoder for binary input data
            self.encoder = MySklearnEncoder.fit()
    
        # List of the feature order the ml model was trained on
        def feature_input_order(self):
            return [...]
    
        # The ML framework the model was trained on
        def backend(self):
            return "pytorch"
    
        # The black-box model object
        def raw_model(self):
            return self._mymodel
    
        # The predict function outputs
        # the continous prediction of the model
        def predict(self, x):
            return self._mymodel.predict(x)
    
        # The predict_proba method outputs
        # the prediction as class probabilities
        def predict_proba(self, x):
            return self._mymodel.predict_proba(x)
    \end{lstlisting}
    \caption{Pseudo-implementation of the \texttt{CARLA} black-box-model wrapper}
    \label{fig:carla_example_model}
\end{figure*}

\subsection{Benchmarking Recourse Methods}
Besides the generation of counterfactual examples, the focus of \texttt{CARLA} lies on benchmarking recourse methods. Users are able to compute evaluation measures to make qualitative statements about usability and applicability.

All measurements, which are described in Section \ref{section:measures}, are implemented in the \textit{Benchmarking} class of \texttt{CARLA} and can be used for every wrapped recourse method. Figure \ref{fig:carla_example_benchmark} shows an example implementation of a benchmarking process based on the variables of Figure \ref{fig:carla_example1}. %This approach can be applied to every use-case stated in Figures \ref{fig:carla_example_data}, \ref{fig:carla_example_model}, \ref{fig:carla_example_recourse}.

\begin{figure*}[h]
    \centering
    \begin{lstlisting}[language=Python]
    
    from carla import Benchmark

    # 1. Initilize the benchmarking class by passing
    # black-box-model, recourse method, and factuals into it
    benchmark = Benchmark(model, gs, factuals)
    
    # 2. Either only compute the distance measures
    distances = benchmark.compute_distances()
    
    # 3. Or run all implemented measurements and create a
    # DataFrame which consists of all results
    results = benchmark.run_benchmark()
    \end{lstlisting}
    \caption{Pseudo-implementation of the \texttt{CARLA} recourse method wrapper}
    \label{fig:carla_example_benchmark}
\end{figure*}

\section{Additional Experimental Results}\label{appendix:add_experiments}
\begin{figure*}[!h]
    \centering
    \includegraphics[scale=0.36]{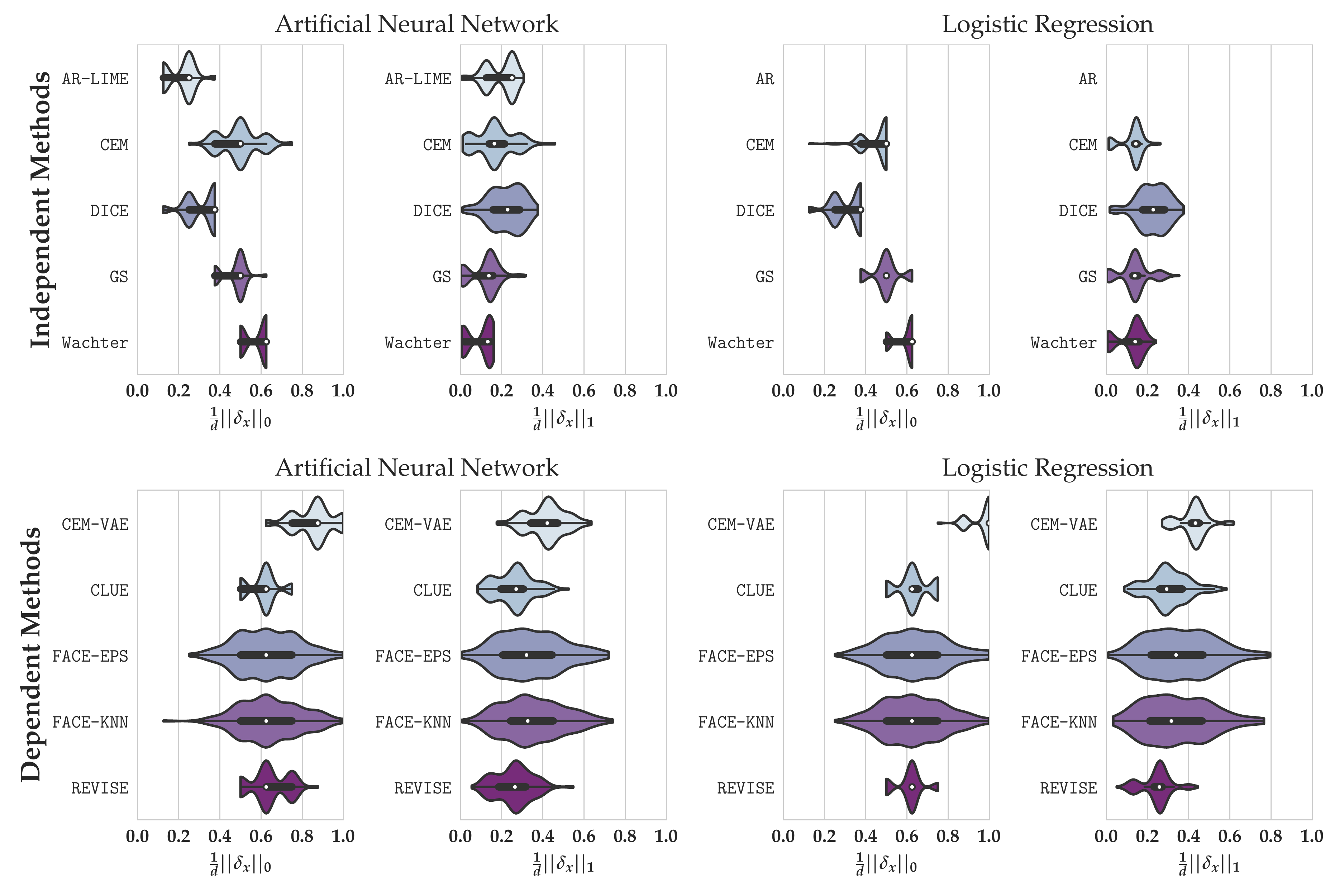}
    \caption{COMPAS Data}
    \vspace{-0.08cm}
    \label{fig:cost_comparison_compas}
\caption{Evaluating the distribution of costs of counterfactual explanations on the COMPAS dataset. For all instances with a negative prediction ($\{x \in \mathcal{D}: f(x) < \theta\}$), we plot the distribution of $\ell_0$ and $\ell_1$ costs of algorithmic recourse as defined in \eqref{eq:costs} for a logistic regression and an artificial neural network classifier. The white dots indicate the medians (lower is better), and the black boxes indicate the interquartile ranges. We distinguish between independence based and dependence based methods.}
\end{figure*}

In this Section, we depict the missing experiments from the COMPAS data set in Figure \ref{fig:cost_comparison_compas} and Table \ref{table:compas_results}. These results underline the trends that we have already highlighted in Section \ref{section:evaluation}.

\begin{table*}[tb]
\centering
\begin{subtable}[b]{\textwidth}

\centering
\resizebox{\columnwidth}{!}{%
\begin{tabular}{ccrrrrrrrrrr}
\toprule
\multicolumn{2}{c}{} & \multicolumn{5}{c}{Artificial Neural Network} & \multicolumn{5}{c}{Logistic Regression} \\
\cmidrule(lr){3-7} \cmidrule(lr){8-12}
Data Set  & Method &   \textit{yNN} &  redund. &  violation &  success &  $\overline{t}(s)$ &   \textit{yNN} &  redund. &  violation &  success &  $\overline{t}(s)$ \\
\cmidrule(lr){1-1} \cmidrule(lr){2-2} \cmidrule(lr){3-7} \cmidrule(lr){8-12}

\multirow{5}{*}{COMPAS} & \texttt{AR(--LIME)} &          \textbf{0.91} &     \textbf{0.00} &       \textbf{0.02} &     0.53 &               0.06 &           -- &      -- &        -- &     0.00 &               \textbf{0.01 }\\
       & \texttt{CEM} &          \textbf{0.98} &     2.29 &       0.43 &     \textbf{1.00 }&               0.89 &          0.93 &     1.88 &       0.99 &     \textbf{1.00} &               0.86 \\
       & \texttt{DICE} &          0.89 &     0.88 &       1.03 &     \textbf{1.00} &               0.09 &          \textbf{0.95 }&     0.94 &       0.90 &     \textbf{1.00} &               0.09 \\
       & \texttt{GS} &          0.44 &     0.97 &       0.03 &     \textbf{1.00} &               \textbf{0.01} &          0.60 &     \textbf{0.64} &       \textbf{0.02} &     \textbf{1.00} &               \textbf{0.01}\\
       & \texttt{Wachter} &          0.56 &     1.77 &       0.74 &     0.66 &              10.90 &          0.50 &     1.21 &       0.79 &     \textbf{1.00} &               0.02 \\
\bottomrule

\end{tabular}
}
\caption{Independence based methods}
\label{tab:compas_independent}
\end{subtable}
\vspace{0.05cm}

\vfill

\begin{subtable}[b]{\textwidth}

\centering
\resizebox{\columnwidth}{!}{%
\begin{tabular}{ccrrrrrrrrrr}
\toprule
\multicolumn{2}{c}{} & \multicolumn{5}{c}{Artificial Neural Network} & \multicolumn{5}{c}{Logistic Regression} \\
\cmidrule(lr){3-7} \cmidrule(lr){8-12}
Data Set  & Method &  \textit{yNN} &  redund. &  violation &  success &  $\overline{t}(s)$ &  \textit{yNN} &  redund. &  violation &  success &  $\overline{t}(s)$ \\
\cmidrule(lr){1-1} \cmidrule(lr){2-2} \cmidrule(lr){3-7} \cmidrule(lr){8-12}

\multirow{5}{*}{COMPAS} & \texttt{CEM--VAE} &          \textbf{1.00 }&     5.59 &       1.98 &     \textbf{1.00} &               0.89 &          1.00 &     6.91 &       2.14 &     \textbf{1.00} &               0.87 \\
       & \texttt{CLUE} &          0.99 &     4.06 &       \textbf{1.08} &     \textbf{1.00} &               2.03 &          \textbf{1.00} &     4.62 &       1.25 &     \textbf{1.00 }&               1.88 \\
       & \texttt{FACE--EPS} &          0.94 &     3.71 &       1.55 &     0.99 &               0.45 &          0.97 &     3.94 &       1.62 &     0.99 &               0.45 \\
       & \texttt{FACE--KNN} &          0.94 &     3.83 &       1.63 &     \textbf{1.00} &               \textbf{0.44} &          0.97 &     3.86 &       1.57 &     \textbf{1.00} &               \textbf{0.44} \\
       & \texttt{REVISE} &          \textbf{1.00} &     \textbf{3.29} &       1.29 &     \textbf{1.00 }&               6.06 &          0.92 &     \textbf{3.15} &       \textbf{1.03} &     \textbf{1.00} &               5.35 \\
\bottomrule

\end{tabular}
}
\caption{Dependence based methods}
\label{tab:compas_dependent}

\end{subtable}

\caption{Summary of COMPAS results for independence and dependence based methods. For all instances with a negative prediction ($\{x \in \mathcal{D}: f(x) < \theta\}$), we compute counterfactual explanations for which we then measure $\text{yNN}$ (higher is better), redundancy (lower is better), violation (lower is better), success rate (higher is better) and time (lower is better). We distinguish between a logistic regression and an artificial neural network classifier. Detailed descriptions of these measures can be found in Section \ref{section:benchmarking_process}. The results are discussed in Appendix \ref{appendix:add_experiments}.}

\label{table:compas_results}
\end{table*}

\section{ML Classifiers} \label{appendix:ml_classifiers}
In this section, we describe how the black--box models $f$ were fitted. \texttt{CARLA} supports different ML libraries to estimate these models (e.g., Pytorch, Tensorflow) as the implementations of the various explanation methods work with a particular ML library. We note that the various explanation methods rely on different binary feature encodings. \texttt{DICE}, for example, requires that binary inputs are supplied as one--hot vectors, while \texttt{FACE} needs binary features encoded in a single column. If this was the case, we fitted two ML models, using the same hyperparameters, and generated CEs with respect to the same set of samples.

To ensure similar behavior between the different ML libraries and encoding variations, each black-box model type has the same structure (e.g., number of hidden layer, number of neurons), and training parameters (e.g., learning rate, epochs, etc.). 

The first model is a multi-layer perceptron, consisting of three hidden layers with 18, 9 and 3 neurons, respectively.  We use ReLu activation functions and binary cross entropy to calculate class probabilities. Optimization of the loss function is done by RMSProp \cite{tieleman2012lecture} using a learning rate of 0.002 for every data set. By performing 25 epochs on COMPAS and 10 epochs on Adult and GMC we reached acceptable performance. Further increasing epochs gave rise to very marginal performance increases. For Adult we use a batch--size of 1024, for COMPAS 25 and for GMC 2048.

To allow a more extensive comparison between CE methods, we choose linear models as the second black--box model category for which we evaluate the CE methods. Again, we optimized these models with RMSProp using a binary cross entropy loss. For Adult, we used 100 epochs and a batch--size of 2048, for COMPAS we choose 25 epochs and batch--size of 128, and for GMC we chose 10 epochs with a batch--size of 2048. The learning rate on every data set is set 0.002. Table \ref{tab:classifier_acc} provides an overview of the model's classification accuracies.

{\color{red}{\begin{table}[H]
\centering
\begin{tabular}{@{}lccc@{}}
\toprule
 & Adult & COMPAS & Give Me Credit \\ \midrule
Logistic Regression & 0.83  & 0.84  & 0.92  \\
Neural Network & 0.84  & 0.85  & 0.93  \\ \bottomrule
\end{tabular}
\caption{Performance of classification models used for generating algortihmic recourse.}
\label{tab:classifier_acc}
\end{table}
}}

\section{COMPAS Data Set Description} \label{appendix:missing_data_set}

The \textbf{COMPAS} data set \cite{Angwin2016} contains data for more than 10,000 criminal defendants in Florida. It is used by the jurisdiction to score defendant's likelihood of reoffending. We kept a small part of the raw data as features like \textit{name}, \textit{id}, \textit{casenumbers} or \emph{date-time} were dropped. The classification task consists of classifying an instance into high risk of recidivism (\textit{score\_text} is high). By converting the feature \textit{race} into \emph{white} and \emph{non-white}, we keep the categorical input binary. Similar to Adult, the immutable features for COMPAS are \textit{age, sex} and \textit{race}.

\section{Hyperparameter Search for the Counterfactual Explanation and Recourse Methods}\label{appendix:hyperparameters}
We generated counterfactual explanations for instances from $H^-$, the set of factuals with negative class predictions.

\paragraph{\textbf{\texttt{AR} and \texttt{AR--LIME}}} It frequently occurred that the action with the lowest cost did not flip the prediction of the black-box classifier. To overcome this problem, we let \texttt{AR} compute a flipset of 150 actions per instance, and subsequently search this set for low--cost CEs. For \texttt{AR--LIME}, we used \texttt{LIME} \citep{ribeiro2016should} and required \texttt{sampling around the instance} to make sure that the coefficients at $x$ were truly local. %The success rate is sensitive to this choice. 

%\paragraph{\textbf{\texttt{AS}}}
%Due to the need for hand-crafted actions, we based our decision for each data set on the importance of its features. The top-3 features with highest \texttt{LIME} importance were used for the actions. 
%Since AS uses search algorithms to find the optimal sequence of actions, we saved computation time by using only 3 actions.
%, and despite of the other CE methods, we generated counterfactuals only for 30 instances.

\paragraph{\textbf{\texttt{CEM}}} After performing grid search, we set the $\ell_1$ weight to 0.9 and the $\ell_2$ weight to 0.1, yielding the strongest performance. For \texttt{CEM-VAE} we set the $\ell_2$ weight to 0.1, and the VAE--weight to 0.9.

\paragraph{\textbf{\texttt{CLUE}}} We use the default hyperparameters from \cite{antoran2020getting}, which are set as a function of the data set dimension $d$. Performing hyperparameter search did not yield results that were improving distances while keeping the same success rate.

\paragraph{\textbf{\texttt{DICE}}}
Since \texttt{DICE} is able to compute a set of counterfactuals for a given instance, we only chose to generate one CE per input instance. We use a \emph{grid search} for the \emph{proximity} and \emph{diversity} weights.

% \paragraph{\textbf{\texttt{EB-CF}}}
% We use the default hyperparameters as described in \cite[Appendix]{mahajan2019preserving}. In general, \texttt{EB--CF} was very sensitive to deviations of these parameters. Deviating too far from the default parameters mostly resulted in success rates of 0. 

\paragraph{\textbf{\texttt{FACE}}}
To determine the strongest hyperparameters for the graph size we conducted a \emph{grid search}. We found that values of $k_{FACE}=50$ gave rise to the best balance of success rate and costs. For the epsilon graph, a radius of 0.25 yields the strongest results to balance between high $\text{yNN}$ and low cost.

\paragraph{\textbf{\texttt{GS}}} We chose 0.02 as the step size with which the sphere is grown. Lower values yield similar results at the costs of higher computational time, while higher values gave worse results.

\paragraph{\textbf{\texttt{REVISE}}}
The \textit{grid search} to find an acceptable learning rate and similarity weight $\lambda$ yielded $\eta=0.1$ and $\lambda = 0.5$ for about 1500 iterations.

\paragraph{\textbf{\texttt{Wachter}}}
For the target loss, we choose the Binary Cross Entropy with a learning rate of $0.01$ and an initial $\lambda$ of $0.01$. For the distance loss, we use the $\ell_1$- norm to measure the similarity between the factual input and the counterfactual point $\CF$.
\end{document}